\documentclass[letterpaper, 10 pt, conference]{ieeeconf}  
\IEEEoverridecommandlockouts
\usepackage{adjustbox}
\usepackage{caption}
\usepackage{cite}

\usepackage{amsmath,amssymb,amsfonts}
\usepackage{graphicx}
\usepackage[linesnumbered,algoruled,boxed,lined]{algorithm2e}
\usepackage{textcomp}
\usepackage{comment}
\usepackage{xcolor}
\usepackage{paralist}
\def\BibTeX{{\rm B\kern-.05em{\sc i\kern-.025em b}\kern-.08em
    T\kern-.1667em\lower.7ex\hbox{E}\kern-.125emX}}
\usepackage{times}  
\usepackage{helvet}  
\usepackage{todonotes}
\usepackage{amsmath}
\usepackage{amssymb}
 
\usepackage{amsthm}
\usepackage{mathtools}
\usepackage{acronym}
\usepackage{courier}  
\newif\ifuseboldmathops
\newif\ifuseittextabbrevs
\useboldmathopstrue   

\ifuseittextabbrevs

\else

\fi

\ifuseboldmathops

\else

\fi

\ifuseboldmathops

\else

\fi

\ifuseboldmathops

\else

\fi

\ifuseboldmathops

	\newcommand{\dom}{\mathop{\bf dom}}   

\else

	\newcommand{\dom}{\mathop{\mathrm{dom}}}   
	
\fi

\renewcommand{\vec}[1]{\mathbf{#1}}

\acrodef{mdp}[MDP]{Markov Decision Process}
\acrodef{pomdp}[POMDP]{Partially Observable Markov Decision Process}

\theoremstyle{definition}

\acrodef{smdp}[Semi-MDP]{Semi-Markov decision process}
\acrodef{rl}[RL]{reinforcement learning}
\acrodef{mcts}[MCTS]{Monte Carlo tree search}
\acrodef{uct}[UCT]{Upper Confidence Bound 1 applied to trees}
\acrodef{scltl}[scLTL]{syntactically co-safe LTL}
\acrodef{ssp}[SSP]{Stochastic Shortest Path}
\acrodef{p2sg}[SG(2)]{Two-player Stochastic Game}
\acrodef{dof}[DOF]{degree of freedom}
\acrodef{cpg}[CPG]{Central Pattern Generator}
\acrodef{nn}[NN]{Neural Network}
\acrodef{snn}[SNN]{Spiking Neural Net}
\acrodef{rstdp}[R-STDP]{Reward-Modulated Spike-Timing-Dependent Plasticity}
\acrodef{gp}[GP]{Genetic Programming}
\acrodef{ppoc}[PPOC]{Proximal Policy Optimization Option-Critics}
\acrodef{dr}[DR]{Domain Randomization}
\acrodef{bibo}[BIBO]{Bounded-input, Bounded-Output}

\acrodefplural{dof}[DOFs]{degrees of freedom}
\acrodefplural{cpg}[CPGs]{Central Pattern Generators}

\usepackage{url}  
\usepackage{graphicx}  
\usepackage{bm}
\usepackage{color}
\usepackage{amsfonts}       
\usepackage{nicefrac}       
\usepackage{microtype}      
\usepackage{lipsum}
\usepackage{amsmath}
\usepackage{subcaption}
\usepackage{algpseudocode} 
\usepackage{tabularx}

\title{\LARGE \bf Learning to Locomote with Deep Neural-Network and CPG-based Control in a Soft Snake Robot
}

\author{Xuan Liu\textsuperscript{1}, Renato Gasoto\textsuperscript{1,2}, Ziyi Jiang\textsuperscript{3}, Cagdas Onal\textsuperscript{1}, Jie Fu\textsuperscript{1}
\thanks{\textsuperscript{1}Xuan Liu is a PhD student under the supervision of Jie Fu and Cagdas Onal in the Robotics Engineering program at Worcester Polytechnic Institute. \textsuperscript{2}Renato Gasoto is affiliated with NVIDIA. \textsuperscript{3}Ziyi Jiang is an undergraduate student in the Electronic Engineering program at Xidian University.} 
\thanks{\tt\small xliu9, rggasoto, cdonal, jfu2@wpi.edu, jzyxbb@outlook.com}%
\thanks{This work was supported in part by the National Science Foundation under grant \#1728412, and by NVIDIA.}%
}

\begin{document}

\maketitle
\thispagestyle{empty}
\pagestyle{empty}

\begin{abstract}
 In this paper, we present a new locomotion control method for soft robot snakes. Inspired by biological snakes, our control architecture is composed of two key modules: A deep reinforcement learning (RL) module for achieving adaptive goal-tracking behaviors with changing goals, and a central pattern generator (CPG) system with Matsuoka oscillators for generating stable and diverse locomotion patterns. The two modules are interconnected into a closed-loop system: The RL module, analogizing the locomotion region located in the midbrain of vertebrate animals, regulates the input to the CPG system given state feedback from the robot. The output of the CPG system is then translated into pressure inputs to pneumatic actuators of the soft snake robot. Based on the fact that the oscillation frequency and wave amplitude of the Matsuoka oscillator can be independently controlled under different time scales, we further adapt the option-critic framework to improve the learning performance measured by optimality and data efficiency. The performance of the proposed controller is experimentally validated with both simulated and real soft snake robots. 
\end{abstract}

\section{Introduction}
Due to their flexible geometric shapes and  deformable materials, soft continuum robots have great potentials in performing tasks under dangerous and cluttered environments \cite{majidi2014soft}. 
However, planning and control of such type of robots remains a challenging problem, as these robots have infinitely many degrees of freedom in their body links, and soft actuators with hard-to-identify dynamics.  

We are motivated to develop an intelligent control framework to achieve serpentine locomotion for goal tracking in a soft snake robot, designed and fabricated by \cite{onal2013autonomous}. %
The crucial observation from nature is that most animals with soft bodies and elastic actuators can learn and adapt to various new motion skills with only a few trials. These mechanisms have been studies for decades, and proposed as \acp{cpg} or neural oscillators. As a special group of neural circuits located in the spinal cord of most animals, \acp{cpg} are able to generate rhythmic and non-rhythmic activities for organ contractions and body movements in animals. Such activities can be activated, modulated and reset by neuronal signals mainly from two directions: bottom-up ascendant feedback information from afferent sensory neurons, or top-down descendant signals from high level modules including mesencephalic locomotor region (MLR) \cite{ijspeert2008central} and motor cortex \cite{roberts1998central, yuste2005cortex}.


\begin{figure}[h!]
    \centering
    \includegraphics[width=0.95\columnwidth]{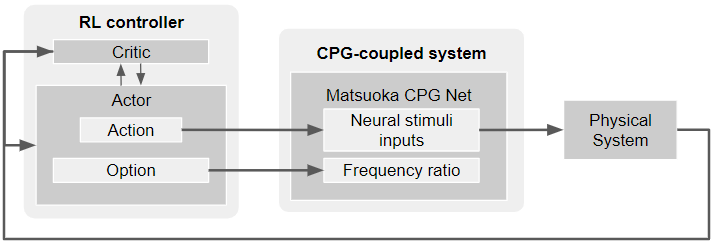}
    \caption{Schematic view of learning based CPG controller.}
    \label{fig:schematic_view}
\end{figure}

In literature, bio-inspired control methods have been studied for the control design of locomotion control of rigid robots, including bipedal \cite{mori2004reinforcement, endo2008learning, nassour2014multi, dzeladini2018cpg} and serpentine locomotion \cite{crespi2005swimming, crespi2008online,ryu2010locomotion,  bing2017towards, wang2017cpg, bing2019end}. The general approach is to generate motion patterns mimicking animals' behaviors and then to track these trajectories with a closed-loop control design. In \cite{ijspeert2008central}, the authors developed a trajectory generator for a rigid salamander robot using Kuramoto \acp{cpg} and used low-level PD controllers to track the desired motion trajectories generated by the oscillator.  In \cite{ryu2010locomotion}, the authors improved the synchronization property of the \ac{cpg} by adapting its frequency parameter with additional linear dynamics. In \cite{wang2017cpg}, the authors introduced a control loop that adjusts the frequency and amplitude of the oscillation for adapting to the changes of the terrain. Another recent work \cite{bing2019end} employed \ac{snn} under the regulation of \ac{rstdp} to map visual information into wave parameters of a phase-amplitude \ac{cpg} net, which generates desired oscillating patterns to locomote a rigid snake robot chasing a red ball. 

Despite the success of bio-inspired control with rigid robots, it may be impossible to expect good performance if we apply the same control scheme to soft robots. This is mainly because that in these approaches, the trajectories generated by \ac{cpg} require high-performance low-level controllers for tracking. The tracking performance cannot be reproduced with soft snake robots due to the nonlinear, delayed and stochastic dynamical response from the soft actuators. In addition, most of these approaches only focus on improving locomotion performance under certain scenario, with a fixed goal position, or limited terrain types. To the best of our knowledge, none of the investigated work can track a randomly generated target with soft snake robot system. 


To this end, we develop a bio-inspired intelligent controller for soft snake robots with two key components: 
 To achieve intelligent and robust goal tracking with changing goals,
 we use model-free reinforcement learning \cite{sutton2000policy,schulman2017proximal} to map the feedback of soft actuators and the goal location, into control commands of a \ac{cpg} network. The \ac{cpg} network consists of coupled Matsuoka oscillators \cite{matsuoka1985sustained}. It acts as a low-level motion controller to generate actuation inputs directly to the soft snake robots for achieving smooth and diverse motion patterns. The two networks forms a variant of cascade control with only one outer-loop, as illustrated in Fig.~\ref{fig:schematic_view}. 
 
To reduce the time and samples required for learning-based control, we leverages dynamic properties of Matsuoka oscillators in designing the interconnection between two networks. 
The \ac{rl} module learns to control \emph{neural stimuli inputs} and \emph{frequency ratio} to a \ac{cpg} network given state feedback from the soft snake robot and the control objective. In analogy to autonomous driving, the neural stimuli inputs play the role of \emph{steering control}, by modulate the amplitudes and phases of the outputs of the \ac{cpg} net in real-time. The frequency ratio plays the role of \emph{velocity control} as the changes the oscillating frequency of the \ac{cpg} net result in changes in the locomotion velocity, measured at the COM of the soft snake robot. 

To conclude, the contributions  include:
\begin{itemize}
    \item A novel bio-inspired tracking  controller for soft snake robots, that combines \emph{robust and optimality} in deep reinforcement learning and \emph{stability and diversity in behavior patterns} in the \ac{cpg} system.
    \item A detailed analysis of Matsuoka oscillators in relation to steering and velocity control of soft snake robots.
    \item Experimental validation in a real soft snake robot.
\end{itemize}

The paper is structured as follows: 
Section~\ref{sec:sys} provides an overview of the robotic system and the state space representation. Section~\ref{sec:CPG} presents the design and configuration of the CPG network. Section~\ref{sec:dis} discusses key properties of the CPG network and the relation to the design of deep neural network for the RL module. Section~\ref{sec:sol} introduces curriculum and reward design for learning goal-reaching locomotion with a soft snake robot. Section~\ref{sec:result} presents the experimental validation and evaluation of the controller in both simulated and real snake robots.

\section{System Overview}
\label{sec:sys}


A full snake robot consists $n$ pneumatically actuated soft links. Each soft link of the robot is made of Ecoflex\texttrademark~00-30 silicone rubber. The links are connected through rigid bodies enclosing the electronic components that are necessary to control the snake robot. In addition, the rigid body components have a pair of one direction wheels to model the anisotropic friction of real snakes. Only one chamber on each link is active (pressurized) at a time. 

\begin{figure}[ht!]
    \centering
    \includegraphics[width=0.8\columnwidth]{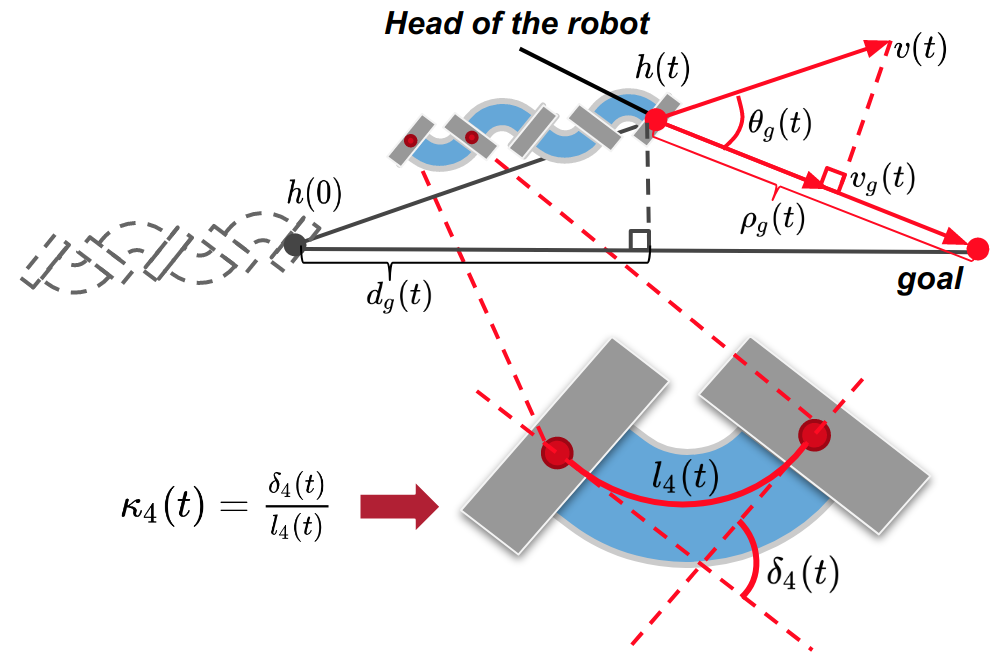}
    \caption{Notation of the state space configuration of the robot.}
    \label{fig:coordinate}
\end{figure}

The configuration of the robot is shown in Figure~\ref{fig:coordinate}. At time $t$, state $h(t) \in \mathbb{R}^2$ is the planar Cartesian position of the snake head, $\rho_g(t) \in \mathbb{R}$ is the distance from $h(t)$ to the goal position, $d_g(t) \in \mathbb{R}$ is the  distance travelled  along the goal direction from the initial head position $h(0)$, $v(t) \in \mathbb{R}$ is the instantaneous planar velocity of the robot, and $v_g(t) \in \mathbb{R}$ is the projection of this velocity vector to the goal direction, $\theta_g(t)$ is the angular deviation between the goal direction and the velocity direction of the snake robot. According to \cite{luo2014theoretical}, the bending curvature of each body link at time $t$ is computed by
$
    \kappa_i(t) = \frac{\delta_i(t)}{l_i(t)}, \text{ for } i=1,\ldots, 4,
$
where $\delta_i(t)$ and $l_i(t)$ are the relative bending angle and the length of the middle line of the $i_{th}$ soft body link.

In \cite{renato2019}, we developed a physics-based simulator that models the inflation and deflation of the air chamber and the resulting deformation of the soft bodies with tetrahedral finite elements. The simulator runs in real-time using GPU. We use the simulator for learning the locomotion controller in the soft snake robot, and then apply the learned controller to the real robot.

\begin{figure*}[ht!]
    \centering
    \vspace{0.75em}
    \includegraphics[width=0.9\textwidth]{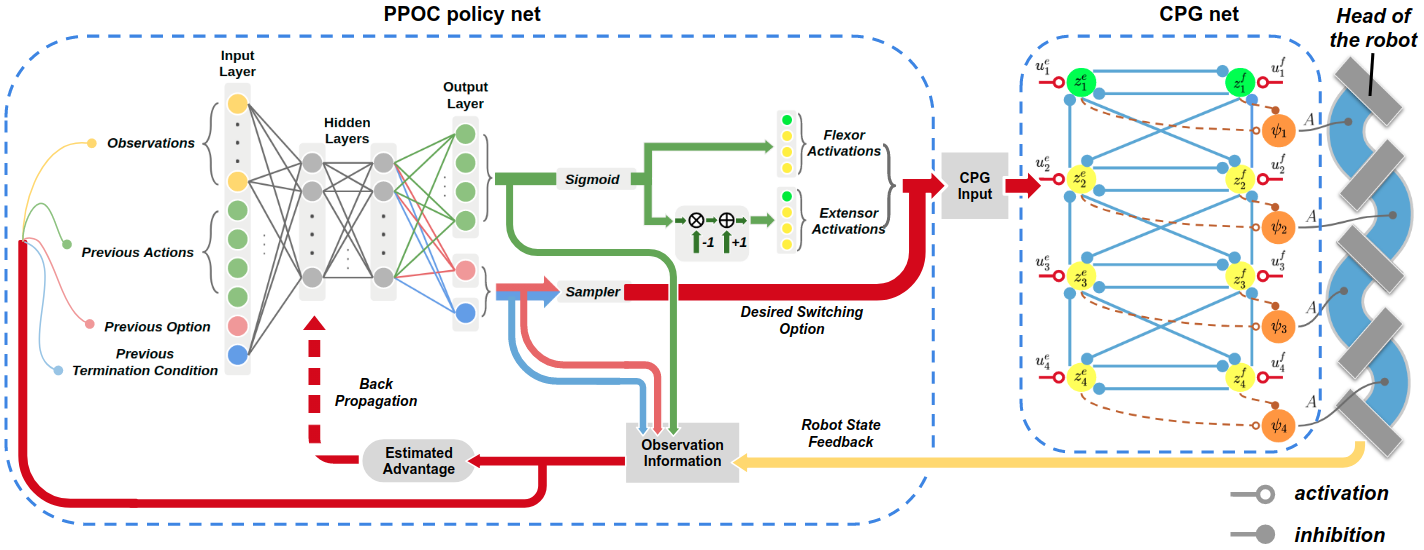}
    \caption{Illustrating the input-output connection of the PPOC-CPG net.}
    \label{fig:rlmatsuoka}
    \vspace{-4ex}
\end{figure*}


\section{Design of CPG Network for a Soft Snake Robot}
\label{sec:CPG}

In this section, we introduce our CPG network design consists of interconnected Matsuoka oscillators \cite{matsuoka1987mechanisms, matsuoka2011analysis}. 

\noindent \textbf{Primitive Matsuoka CPG: }A primitive Matsuoka CPG consists a pair of mutually inhibited neuron models. The dynanmical model of a primitive Matsuoka CPG is given as follows,
\begin{align}
\label{eq:matsuoka}
    &K_f \tau_r \Dot{x}_i^e = -x_i^e - a z_i^f - b y_i^e - \sum_{j=1}^N w_{ji}y_j^e + u_i^e\\ \nonumber
    &K_f \tau_a \Dot{y}_i^e = z_i^e - y_i^e\\ \nonumber
    &K_f \tau_r \Dot{x}_i^f = -x_i^f - a z_i^e - b y_i^f - \sum_{j=1}^N w_{ji}y_j^f + u_i^f\\ \nonumber
    &K_f \tau_a \Dot{y}_i^f = z_i^f - y_i^f\\ \nonumber
    &\psi_i = A (z_i^e - z_i^f), 
\end{align}
where the subscripts $e$ and $f$  represent variables related to extensor neuron and flexor neuron, respectively, the tuple $(x_i^q, y_i^q)$, $q \in  \{e,f\}$ represents the activation state and self-inhibitory state of $i$-th neuron respectively,  $z_i^q = g(x_i^q) = \max(0, x_i^q)  $ is the output of $i$-th neuron,  $b \in \mathbb{R}$ is a weight parameter,   $u_i^e, u_i^f$ are the tonic inputs to the oscillator, and  $K_f \in \mathbb{R}$ is the frequency ratio.
The set of parameters in the system includes: the discharge rate $\tau_r \in \mathbb{R}$, the adaptation rate $\tau_a \in \mathbb{R}$,  the mutual inhibition weights between flexor and extensor $a\in \mathbb{R}$ and the inhibition weight $w_{ji}\in \mathbb{R}$ representing the coupling strength with neighboring primitive oscillator. In our system, all coupled signals including $x_i^q, y_i^q$ and $z_i^q$ ($q \in  \{e,f\}$) are inhibiting signals (negatively weighted), and only the tonic inputs are activating signals (positively weighted).

\noindent \textbf{Configurating the Matsuoka CPG network: }The structure of the proposed CPG network is shown in Fig.~\ref{fig:rlmatsuoka}. The network includes four linearly coupled primitive Matsuoka oscillators. 
It is an inverted, double-sized version of Network VIII introduced in Matsuoka's paper \cite{matsuoka1987mechanisms}. The network includes four pairs of nodes. Each pair of nodes (e.g., the two nodes colored green/yellow) in a row represents a primitive Matsuoka CPG \eqref{eq:matsuoka}. The edges are correspond to the coupling relations among the nodes. In this graph, all the edges with hollowed endpoints are positive activating signals, while the others with solid endpoints are negative inhibiting signals. The oscillators are numbered $1 \text{ to } 4$ from head to tail of the robot. 

The outputs $\bm{\psi} = [\psi_1,\psi_2,\psi_3,\psi_4 ]^T$ from the primitive oscillators are the normalized input ratio in the real region $[-1, 1]$. We let $\psi_i = 1$ for the full inflation of the $i$-th extensor actuator and zero inflation of the $i$-th flexor actuator, and vice versa for $\psi_i = -1$. The actual pressure input to the $i$-th chamber is $\lambda_i\cdot\psi_i$, where $\lambda_i$ is the maximal pressure input of each actuator. The primitive oscillator with green nodes controls the oscillation of the head joint. This head oscillator also contributes as a rhythm initiator in the oscillating system, followed by the rest parts oscillating with different phase delays in sequence. Figure~\ref{fig:rlmatsuoka} shows all activating signals to the CPG network. For simplicity, we'll use a vector \begin{equation}
\label{eq:bvec}
    \bm{u} = [u_1^e, u_1^f, u_2^e, u_2^f, u_3^e, u_3^f, u_4^e, u_4^f]^T
\end{equation} to represent all tonic inputs to the \ac{cpg} net.
To achieve stable and synchronized oscillations of the whole system, the following constraint must be satisfied \cite{matsuoka1985sustained}:
\begin{align}
\label{eq:oscilexist}
    (\tau_a-\tau_r)^2 < 4\tau_r\tau_a b,
\end{align}
where $\tau_a, \tau_r, b > 0$. To satisfy this constraint, we can set the value of $b$ much greater than both $\tau_r$ and $\tau_a$, or make the absolute difference $|\tau_r-\tau_a|$ sufficiently small. To determine the hyper-parameters in the \ac{cpg} network that generate more efficient locomotion pattern, we employed a \ac{gp} algorithm similar to \cite{ijspeert1999evolving} before integrating it with the proposed NN controller. In this step, all tonic inputs are set as constant integer $1$ for the simplicity of fitness evaluation. 


\begin{figure*}[h!]
\centering
\vspace{0.75em}
\begin{subfigure}[b]{0.09\textwidth}
\includegraphics[width=0.99\columnwidth]{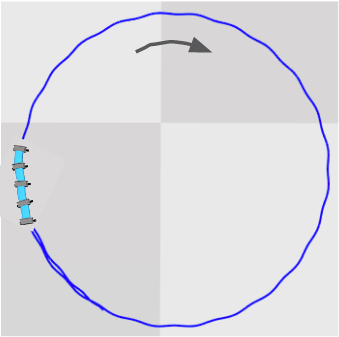}\caption{ \label{fig:offset}}
\end{subfigure}
\hspace{-0.01\textwidth}
\begin{subfigure}[b]{0.09\textwidth}
\includegraphics[width=0.99\columnwidth]{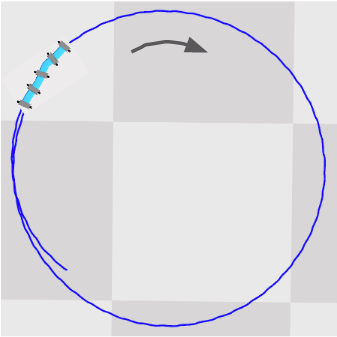}
\caption{ 
\label{fig:duty}}
\end{subfigure}
\begin{subfigure}[b]{0.37\textwidth}
\includegraphics[width=\columnwidth]{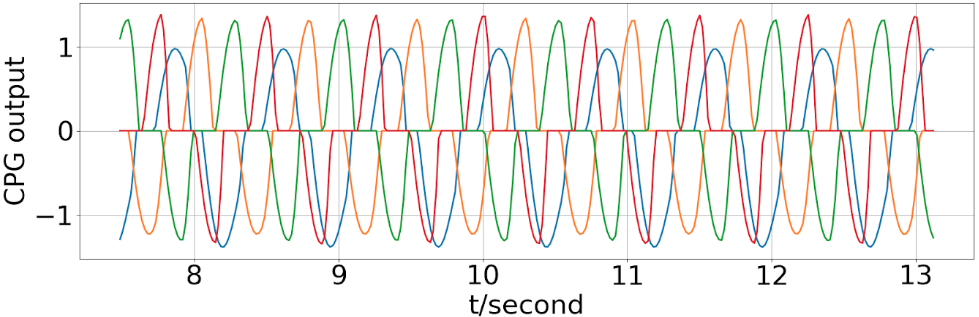}
\caption{
\label{fig:offsetwave}}
\end{subfigure}
\begin{subfigure}[b]{0.42\textwidth}
\includegraphics[width=\columnwidth]{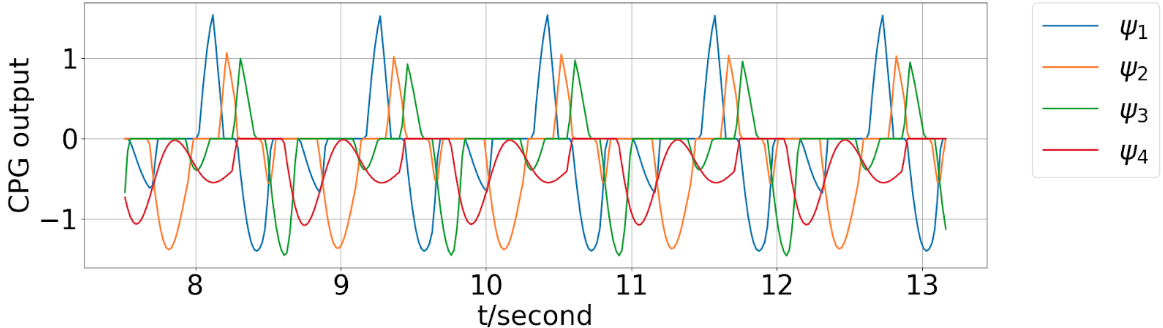}
\caption{
\label{fig:dutywave}}
\end{subfigure}
\caption{ (a) Locomotion trajectory with biased amplitudes of tonic inputs $\bm{u}=[0.4, 0.6, 0.5, 0.5, 0.5, 0.5, 0.5, 0.5]^T$ \eqref{eq:bvec} to the CPG net. (b) Locomotion trajectory with biased duty cycles of tonic inputs. And plots of CPG outputs corresponding to tonic inputs with (c) biased amplitudes and (d) biased duty cycles.} 
\vspace{-4ex}
\label{fig:turn}
\end{figure*}

We define the fitness function--the optimization criteria--in \ac{gp} as $F(t) = a_1| v_d(t)| - a_2 |\theta_d(t)| + a_3 |s_d(t)|),$
where all coefficients $a_1, a_2, a_3,T \in \mathbb{R}^+$ are constants \footnote{In experiments, the following  parameters are used: $a_1= 40.0$, $a_2= 100.0$, $a_3= 50.0$, and $T=6.4 \text{ sec}$. }. 
This fitness function is a weighted sum over the robot's instantaneous velocity, angular deviation, and total traveled distance on a fixed straight line at termination time $t=T$. 
A better fitted configuration is supposed to maintain oscillating locomotion after a given period of time $T$, with faster locomotion speed $|v_d(T)|$ along the original heading direction. In addition, the locomotion pattern is also required to have less angular deviation from the initial heading direction (with a small $ |\theta_d(T)|$), and with overall a longer travelled distance along the initial direction ($|s_d(T)|$). 


The desired parameter configuration found by \ac{gp} is given by Table.~\ref{tab:config} in the Appendix. 



\section{Design of Learning Based Controller with Matsuoka CPG Network}
\label{sec:dis}
With constant tonic inputs, the designed Matsuoka \ac{cpg} net can generate stable symmetric oscillations to efficiently drive the soft snake robot slithering straight forward, it does not control the steering and velocity to achieve goal-reaching behaviors with potentially time-varying goals. 
Towards intelligent navigation control, we employ a model-free RL algorithm, proximal policy optimization (PPO) \cite{schulman2017proximal}, as the `midbrain' of the \ac{cpg} network. The algorithm is to learn the optimal policy that takes state feedback from the robot and control tonic inputs and frequency ratio of the \ac{cpg} net to generate proper oscillating waveform for reaching a goal. 
We would like to reduce the data and computation required for the learning to converge by leveraging crucial features of the Matsuoka \ac{cpg}.
Next, we present our configuration of RL guided by dynamical properties of Matsuoka \ac{cpg} net.


\subsection{Encoding tonic inputs for steering control}
\label{sec:steer}

Steering and velocity control are key to goal-directed locomotion. 
Existing methods realize steering by either directly adding a displacement \cite{ijspeert2008central} to the output of the \ac{cpg} system, or using a secondary system such as a deep neural network to manipulate the output from multiple \ac{cpg} systems \cite{mori2004reinforcement}.
We show that the maneuverability of Matsuoka oscillator provides us a different approach--tuning tonic inputs to realize the desired wave properties for steering.



In this subsection, we show two different combinations of tonic inputs capable of generating imbalanced output trajectories that result in steering of the robot. One way is to apply biased values to each pair of tonic inputs. Another way is to introduce different duty cycles between the actuation of extensor and flexor. 

\begin{figure}[h!]
    \centering
    \includegraphics[width=0.9\columnwidth]{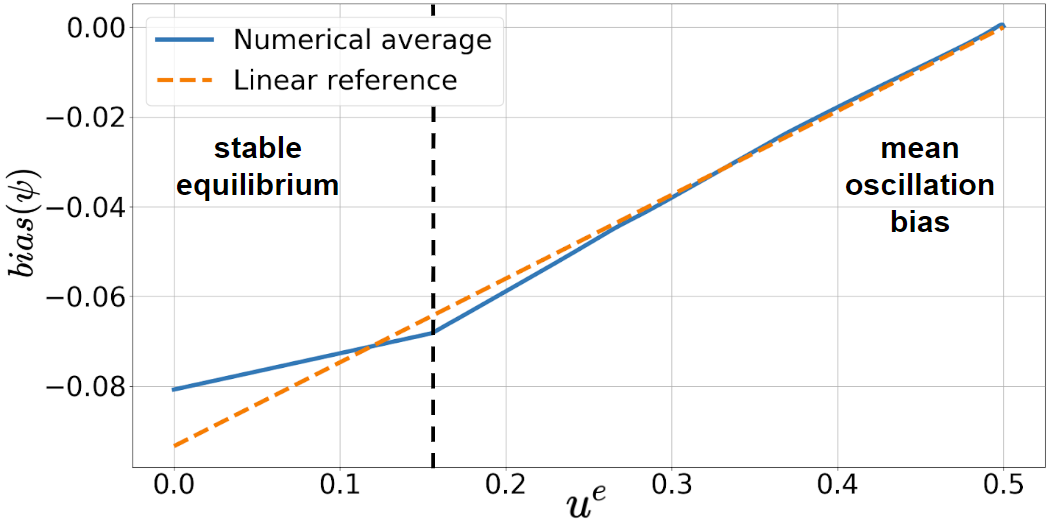}
    \caption{Relation between oscillation bias and activation amplitude of $u^e$ from a single primitive oscillator when $u^e+u^f=1$.}
    \label{fig:bias}
    \vspace{-2ex}
\end{figure}



Figure~\ref{fig:bias} shows the bias in the output $\psi_i$ given the input $u^e_i$ changing from $0$ to $0.5$ and $u^f_i = 1-u^e_i$ in a primitive Matsuoka oscillator. 
Here, we approximate the bias of the steady-state output oscillation trajectory by taking the time-average of the
 trajectory \footnote{Based on Fourier series analysis, given a continuous real-valued $P$-periodic function $\psi_i(t)$, the  constant  term of its Fourier series has the form
$    \frac{1}{P}\int_P \psi_i(t) dt.
$}.
From Fig.~\ref{fig:bias}, we observe a linear mapping between tonic input $u^e$ of a primitive oscillator and the bias of output, given $u^f +u^e=1$. 


In other words, the steering bias of a primitive Matsuoka oscillator is proportional to the amplitude of $u^e$ when $u^e$ and $u^f$ are exclusive within $[0, 1]$. This key observation allows us to introduce a dimension reduction on the input space of the \ac{cpg} net: Instead of controlling $u^e_i, u^f_i$ for $i=1,\ldots, n$ for $n$-link snake robot, we only need to control $u^e_i$ for $i=1,\ldots, n$ and let  $u^f_i=1-u^e_i$.
As  the  tonic inputs have to be positive in Matsuoka oscillators,  we define a four dimensional action vector $\vec{a}= [a_1, a_2, a_3, a_4]^T \in \mathbb{R}^4$ and map $\vec{a}$ to tonic input vector $\vec{u}$ as follows,
\begin{equation}
\label{eq:decoder}
u_i^e = \frac{1}{1+e^{-a_i}},\text{ and } 
               u_i^f =1-u_i^e, \text{ for } i=1,\ldots, 4.
\end{equation}
 This mapping bounds the tonic input within $[0, 1]$. 
The dimension reduction enables more efficient policy search in \ac{rl}. Furthermore, different action vector $\vec{a}$ can be chosen to stabilize the system to limit cycles or equilibrium points.



Under the constraint of Eq.~\eqref{eq:decoder}, steering can also be achieved by switching between different tonic input vectors  periodically. When the duty cycle of actuating signals are different between flexors and extensors, an asymmetric oscillating pattern will occur. We construct two tonic input vectors $\bm{u_1}$ and $\bm{u_2}$, with $\bm{u_1} = [1, 0, 1, 0, 1, 0, 1, 0]^T$ and $\bm{u_2} = [0, 1, 0, 1, 0, 1, 0, 1]^T$. As Fig.~\ref{fig:dutywave} shows, when we set the duty cycle of $\bm{u_1}$ to be $1/12$ in one oscillating period, the rest $11/12$ of time slot for will be filled with $\bm{u_2}$. The \ac{cpg} output on each link shows an imbalanced oscillation with longer time duration on the negative amplitude axis, indicating longer bending time on the flexor. As a result, the robot makes a clockwise (right) turn, with a circle trajectory presented in Fig.~\ref{fig:duty}. 

We compare the steering control with two different methods in Fig.~\ref{fig:offset} and ~\ref{fig:duty}. The corresponding trajectories in joint space are shown in Fig.~\ref{fig:offsetwave} with biased amplitudes of  tonic inputs and Fig.~\ref{fig:dutywave} with biased duty cycle of tonic inputs. In both experiments, the \ac{cpg} outputs present noticeable biases to the negative amplitude axis. This indicates that all of the soft actuators are bending more to the right-hand side of the robot during the locomotion.



\subsection{Frequency modulation for velocity control}
\begin{figure}[h!]
    \centering
    \includegraphics[width=0.9\columnwidth]{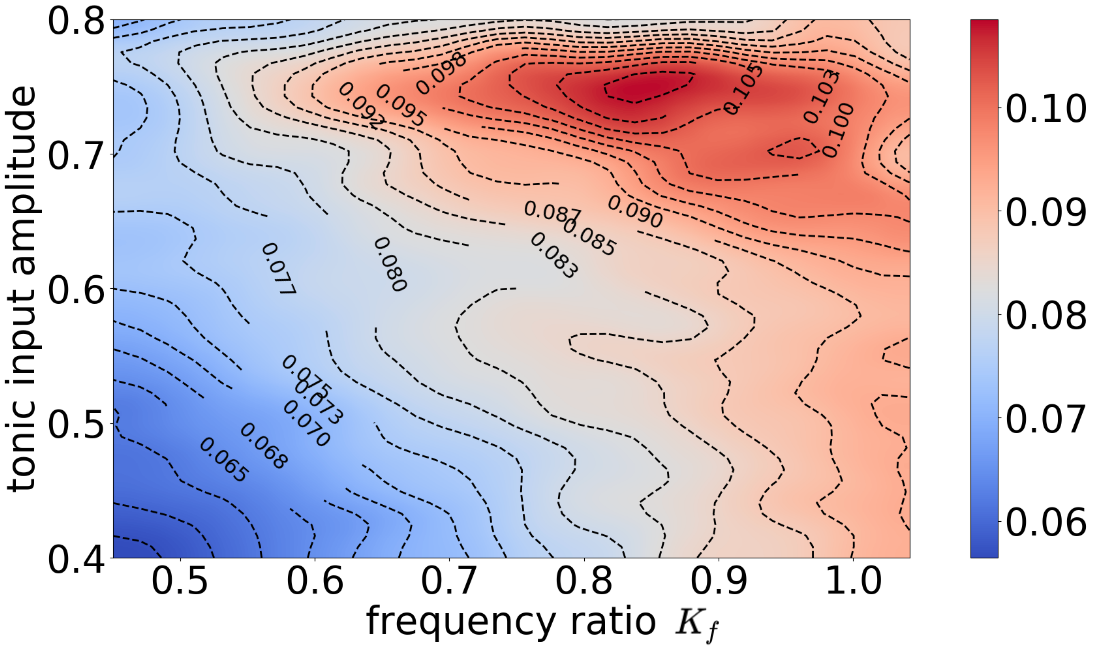}
    \caption{Relating oscillating frequency and amplitude to the average linear velocity of serpentine locomotion.}
    \label{fig:ampfreqvel}
    \vspace{-2ex}
\end{figure}

As seen in Eq.~\eqref{eq:decoder}, the  constraint on each pair of tonic inputs prevents us from controlling locomotion speed with different amplitudes tonic inputs. We would like to determine another input to \ac{cpg} net that controls the locomotion velocity.   

Based on \cite[Eq. (5), Eq. (6)]{matsuoka2011analysis}, since $K_f$ and $\vec{u}$ (Eq.~\eqref{eq:bvec}) are the control inputs, the mapping relations can be concluded as
\begin{align}
\label{eq:kf_property}
      \hat{\omega} \propto \frac{1}{\sqrt{K_f}},  ~\forall~ q\in\{e, f\}, i \in \{1, 2, 3, 4\},
\end{align}
where $\hat{\omega}$ is the natural frequency of the oscillator.

Besides, the oscillation amplitude $\hat{A}$ is linearly proportional to the amplitude of tonic inputs, that is, 
\begin{equation}
    \label{eq:amplitude}
    \hat{A} \propto A(u_i^q),  ~\forall~ q\in\{e, f\}, i \in \{1, 2, 3, 4\},
\end{equation}
where $A(\cdot)$ is the amplitude function of a rhythmic signal.

Equations \eqref{eq:kf_property} and \eqref{eq:amplitude} show that the frequency and amplitude of the Matsuoka \ac{cpg} system can be controlled \emph{ independently} by the frequency ratio $K_f$ and the tonic inputs $u_i^q$, for $q \in \{e, f\}$. 
Figure~\ref{fig:ampfreqvel} shows the distribution of locomotion velocity over different amplitudes and frequencies by taking $2500$ uniform samples within the region $u_i^q \in [0.4, 0.8]$ for $q\in \{e, f\}, i \in \{1, 2, 3, 4\}$ with all $u_i^q$ to be the same in one sample, and $K_f \in [0.45, 1.05]$. Noticed that in Fig.~\ref{fig:ampfreqvel}, with fixed tonic input, the average velocity increase nearly monotonically with the frequency ratio $K_f$. While the amplitude of tonic input does not affect the velocity that much, especially when $K_f$ is low. Given this analysis, we use $K_f$ to control the velocity of the robot.

It is noted that the frequency ratio $K_f$ only influences the strength but not the direction of the vector field of the Matsuoka \ac{cpg} system. Thus, manipulating $K_f$ will not affect the stability of the whole \ac{cpg} system. 




\subsection{The NN controller}
We have now determined the encoded input vector of the \ac{cpg} net to be vector $\vec{a}$ and frequency ratio $K_f$. This input vector of the \ac{cpg} is the output vector of the NN controller. The input to the NN controller is the state feedback of the robot, given by $s=[\rho_g, \Dot{\rho_g}, \theta_g, \Dot{\theta_g}, \kappa_1, \kappa_2, \kappa_3, \kappa_4]^T \in \mathbb{R}^8$ (see Fig.~\ref{fig:coordinate}). Next, we present the design of the NN controller.

We adopt a hierarchical reinforcement learning method called the option framework \cite{sutton1999between,precup2000temporal} to learn the optimal controller regulating the tonic inputs (low-level primitive actions) and frequency ratio (high-level options) of the \ac{cpg} net. The low-level primitive actions are computed at every time step. The high-level option changes infrequently as the robot needs not to change velocity very often for smooth locomotion. 
Specifically, each option is defined by $\langle {\cal{I}}, \pi_y: S\rightarrow \{y\} \times \dom(\vec{a}), \beta_{y} \rangle $ where ${\cal I}= S$ is a set of initial states. By letting ${\cal I}=S$, we  allow the frequency ratio to be changed at any state in the system. Variable $y\in \dom(K_f)$ is a value of frequency ratio, and $\beta_y: S\rightarrow[0,1]$ is the termination function such that $\beta_y(s)$ is the probability of changing from the current frequency ratio to another frequency ratio. 




The options share  the same \ac{nn} for their  intro-option policies and the same \ac{nn} for termination functions. However, these \acp{nn} for intro-option policies take different frequency ratios. The set of parameters to be learned by policy search include parameters for intra-option policy function approximation, parameters for termination function approximation, and parameters for high-level  policy function approximation (for determining the next option/frequency ratio). \ac{ppoc} in the openAI Baselines \cite{baselines}  is employed as the policy search in the RL module.

Let's now review the control architecture in Figure~\ref{fig:rlmatsuoka}. We have a  Multi-layer perceptron (MLP) neural network with two hidden layers to approximate the optimal control policy that controls the inputs of the \ac{cpg} net in Eq.~\eqref{eq:matsuoka}. The output layer of MLP is composed of action $\vec{a}$ (green nodes), option in terms of frequency ratio (pink node) and the terminating probability (blue node) for that option. The input of NN consists of state vector (yellow nodes) and its output from the last time step. The purpose of this design is to let the actor network learn the unknown dynamics of the system by tracking the past actions in one or multiple steps \cite{mori2004reinforcement, peng2018sim, hwangbo2019learning}.
Given the BIBO stability of the Matsuoka \ac{cpg} net \cite{matsuoka1985sustained} and that of the soft snake robots, we ensure that the closed-loop robot system with the PPOC-CPG controller is BIBO stable. Combining with Eq.~\eqref{eq:decoder} that enforces limited range for all tonic inputs, this control scheme is guaranteed to work at bounded status.

\section{Learning-based Goal Tracking Control Design}
\label{sec:sol}

In this section, we introduce the design of curriculum and reward function for learning goal-tracking behaviors with the proposed controller. 
\subsection{Task curriculum}
\begin{figure}[h!]
    \centering
    \vspace{0.75em}
    \includegraphics[width=0.9\columnwidth]{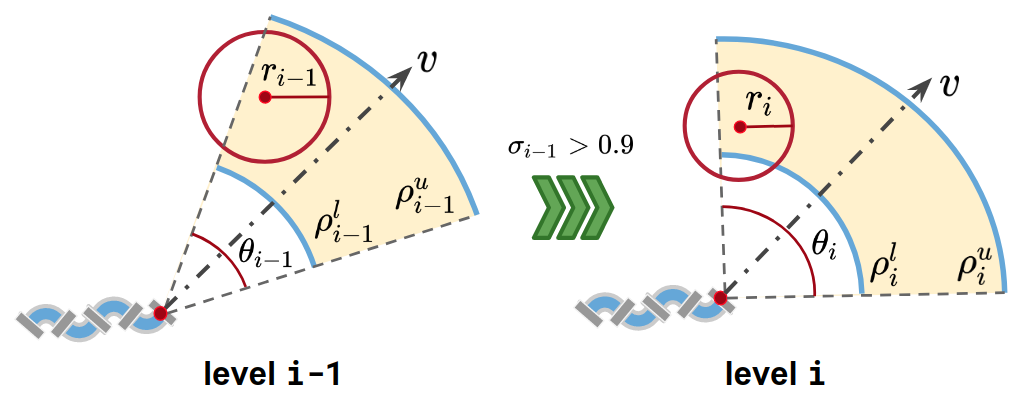}
    \caption{Task difficulty upgrade from level $i-1$ to level $i$. As the curriculum level increases, goals are sampled at a narrower distance and wider angle, and acceptance area gets smaller.}
    \label{fig:curriculumlvlup}
    \vspace{-2ex}
\end{figure} 
 Curriculum teaching \cite{karpathy2012curriculum} is used to accelerate motor skills learning under complex goal reaching task scenario. Through starting the trials with easy-to-reach goals, the agent will learn good policies more quickly. As the level increases, the robot improves the policy with more complex tasks.
We design different levels of tasks as follows: 
At task level $i$, the center of goal is sampled from the 2D workspace based on the current location and head direction of the robot. For each sampled goal, we say the robot reaches the goal if its is $r_{i}$ distance away from the goal. The sampling distribution is a uniform distribution in the fan area determined by the range of angle $\theta_i$ and distance bound $[\rho_i^l, \rho_i^u]$ in polar coordinate given by the predefined curriculum.

As shown in Fig.~\ref{fig:curriculumlvlup}, when the task level increases, we have 
$r_{i}<r_{i-1}$, $\theta_i> \theta_{i-1}$,  $\rho_i^u >\rho_{i-1}^{u}$, and $\rho_i^u-\rho_i^l < \rho_{i-1}^u-\rho_{i-1}^l$; that is, the robot has to be closer to the goal in order to success and receive a reward, the goal is sampled in a range further from the initial position of the robot.  We select discrete sets of $\{r_i\}, \{\theta_i\}$, $[\rho_i^{l},\rho_i^{u}]$ and determine a curriculum table. We train the robot in simulation starting from level $0$. The task level will be increased to level $i+1$ from level $i$ if the controller reaches the desired success rate $\sigma_i$, for example, $\sigma_i=0.9$ indicates at least $90$ successful goal-reaching tasks out of $n=100$ at level $i$. 

\subsection{Reward design}
Based on the definition of goal-reaching tasks and their corresponding level setups, the reward is defined as 
\[
R(v_g,\theta_g) = c_v|v_g| + c_g \cos(\theta_g(t))\sum_{k=0}^i{\frac{1}{r_k} I(l_g(t) < r_k)},
\]
where $c_v, c_g \in \mathbb{R}^+$ are constant weights, $v_g$ is the velocity towards the goal, $\theta_g$ is the angular deviation from the center of goal, $r_k$ defines the goal range in task level $k$, for $k=0,\ldots, i$, $l_g$ is the linear distance between the head of the robot and the goal, and $I(l_g(t) < r_k)$ is an indicator function that outputs one if the robot head is within the goal range for task level $k$. 

This reward trades off two objectives. The first term, weighted by $c_v$, encourages movement toward the goal. The second term, weighted by $c_g$, rewards the learner given the level of curriculum the learner has achieved for the goal-reaching task. For every task, if the robot enters the goal range in task level $i$, it will receive a summation of rewards $1/r_k$ for all $k\le i$ (the closer to the goal the higher this summation), shaped by the approaching angle $\theta_g$ (the closer the angle to zero, the higher the reward).

If the agent reaches the goal defined by the current task level, a new goal is randomly sampled in the current or next level (if the current level is completed). 
There are two failing situations, where the desired goal will be re-sampled and updated. The first situation is starving, which happens when the robot stops moving for a certain amount of time. The second case is missing the goal, which happens when the robot keeps moving towards the wrong direction to the goal region for a certain amount of time.




\section{Experiment validation}
\label{sec:result}


 We used a four-layered neural net configuration with $128\times128$ hidden layer neurons. 
At every step, the algorithm samples the current termination function to decide whether to terminate the current option and obtain a new frequency ratio $K_f$ or keep the previous option. 
The backpropagation of the critic net was done with Adam Optimizer and a step size of $5e-4$. The starvation time for failing condition is $60 \text{ ms}$. The missing goal criterion is determined by whenever $v_g(t)$ stays negative for over $30$ time steps.


\subsection{Policy training}

\begin{figure}[h!]
\centering
\vspace{0.75em}
\begin{subfigure}[b]{0.9\columnwidth} 
\includegraphics[width=0.9\columnwidth]{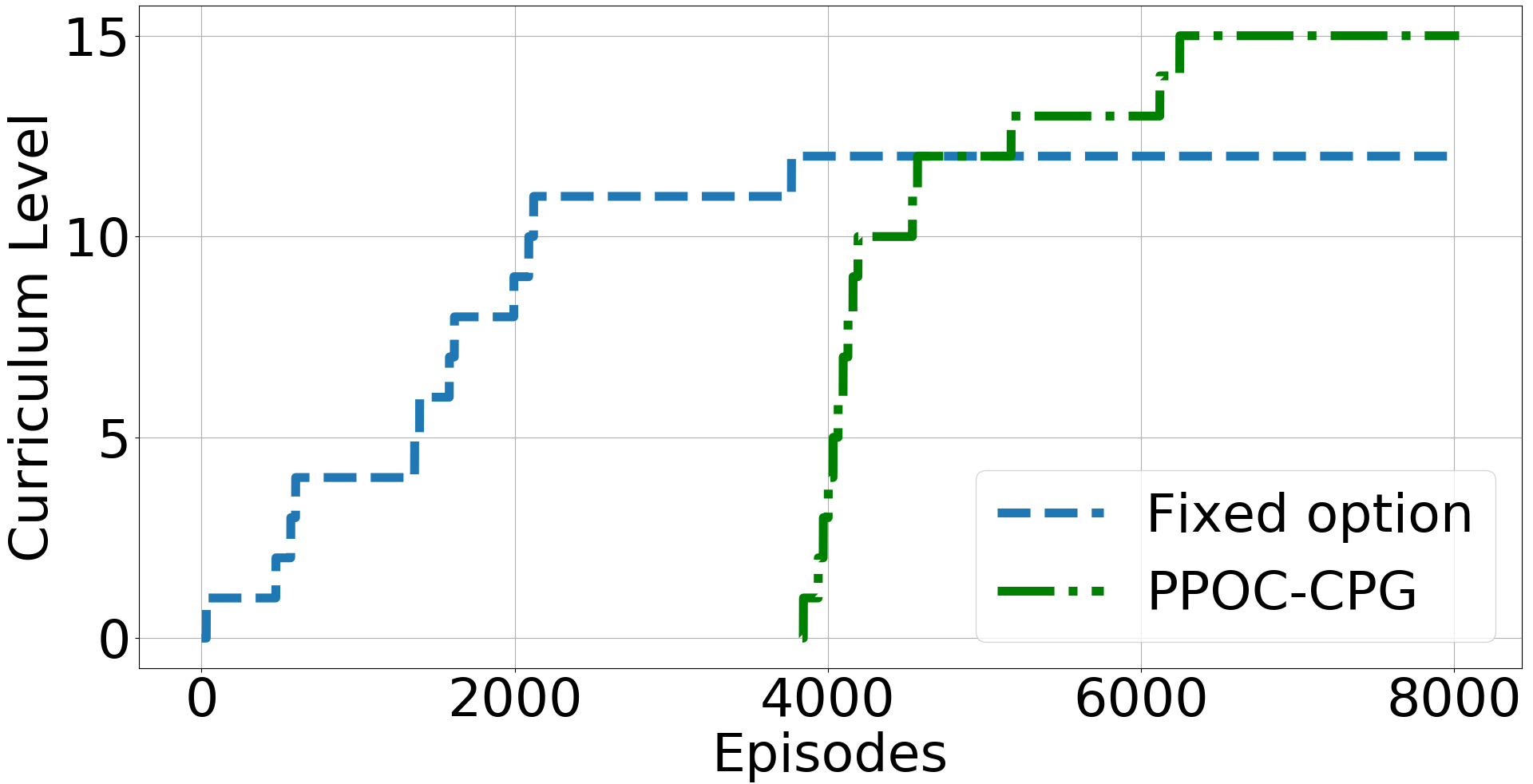} 
\caption{
\label{fig:simlvl}}
\end{subfigure}
\hspace{-0.02\textwidth}
\begin{subfigure}[b]{0.9\columnwidth} 
\includegraphics[width=0.9\columnwidth]{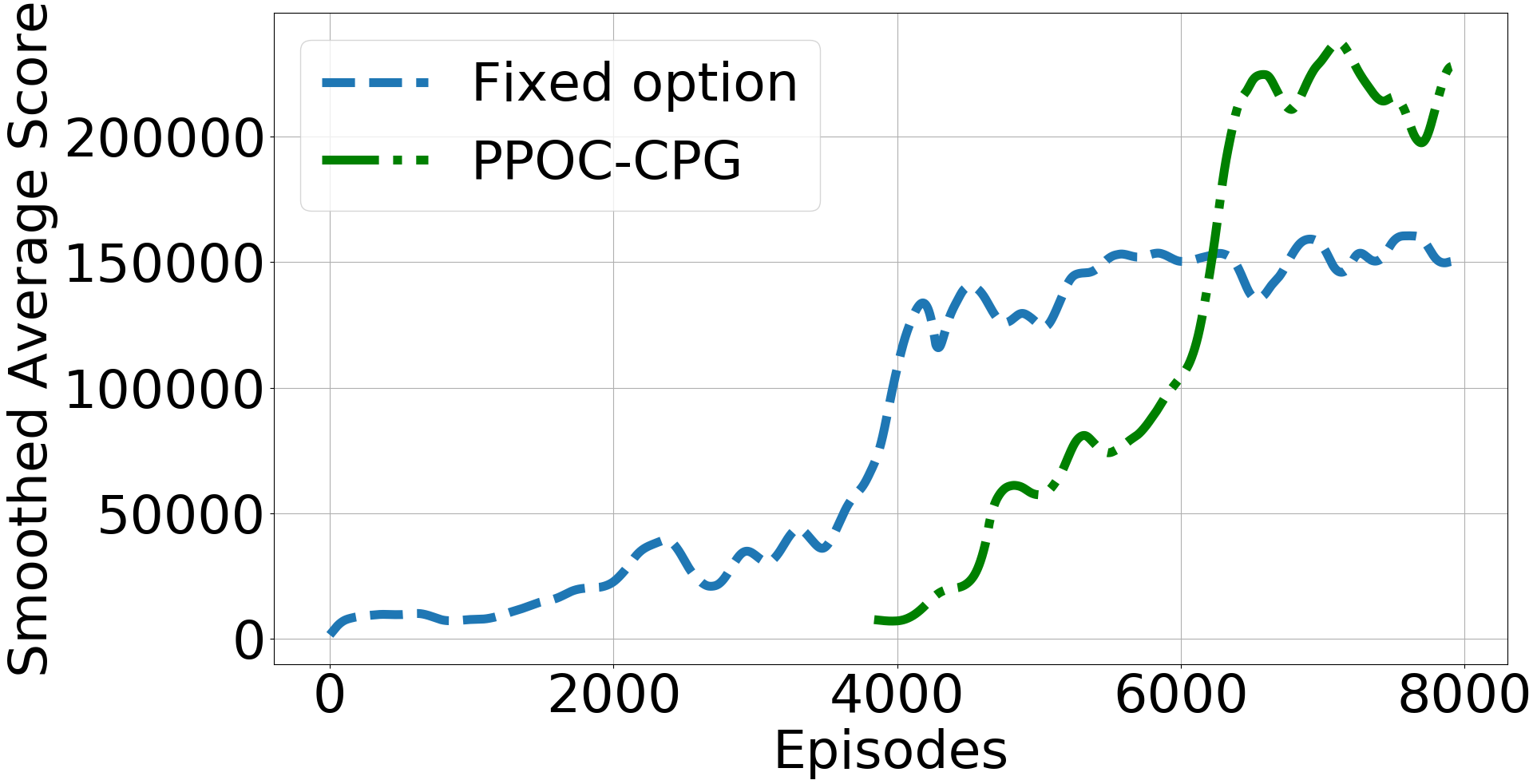}
\caption{
\label{fig:simrew}}
\end{subfigure}
\caption{ (a) Learning progress of task level and (b) average learning score in the goal-reaching task.}
\vspace{-2ex}
\label{fig:trainresult}
\end{figure}
 
Figure~\ref{fig:simlvl} shows improving performance over different levels versus the number of learning episodes. Figure \ref{fig:simrew} shows the corresponding total reward with respect to the number of learning episodes. As shown in Fig.~\ref{fig:simlvl}, we first train the policy net with fixed options (at this moment, the termination probability is always $0$, and a fixed frequency ratio $K_f= 1.0$ is used). When both the task level and the reward cannot increase anymore (at about 3857 episodes), we  allow the learning algorithm to change $K_f$ along with termination function $\beta$, and keep training the   policy until the highest level in the curriculum is passed. In this experiment, the learning algorithm equipped with stochastic gradient descent converges to a near-optimal policy after $6400$ episodes of training. The whole process takes about $12$ hours with $4$ snakes training in parallel on a workstation equipped with an Intel Core i7 5820K, 32GB of RAM, and one NVIDIA GTX1080 ti GPU.

In order to compensate for simulation inaccuracies, most notably friction coefficients, we employed a domain randomization technique \cite{tobin2017domain}, in which a subset of physical parameters are sampled from a distribution with mean on the measured value. The \ac{dr} parameters used for training are on Table \ref{tab:dr}, on the Appendix.



\begin{figure}[h!]
    \centering
    \vspace{2ex}
    \includegraphics[width=0.95\columnwidth]{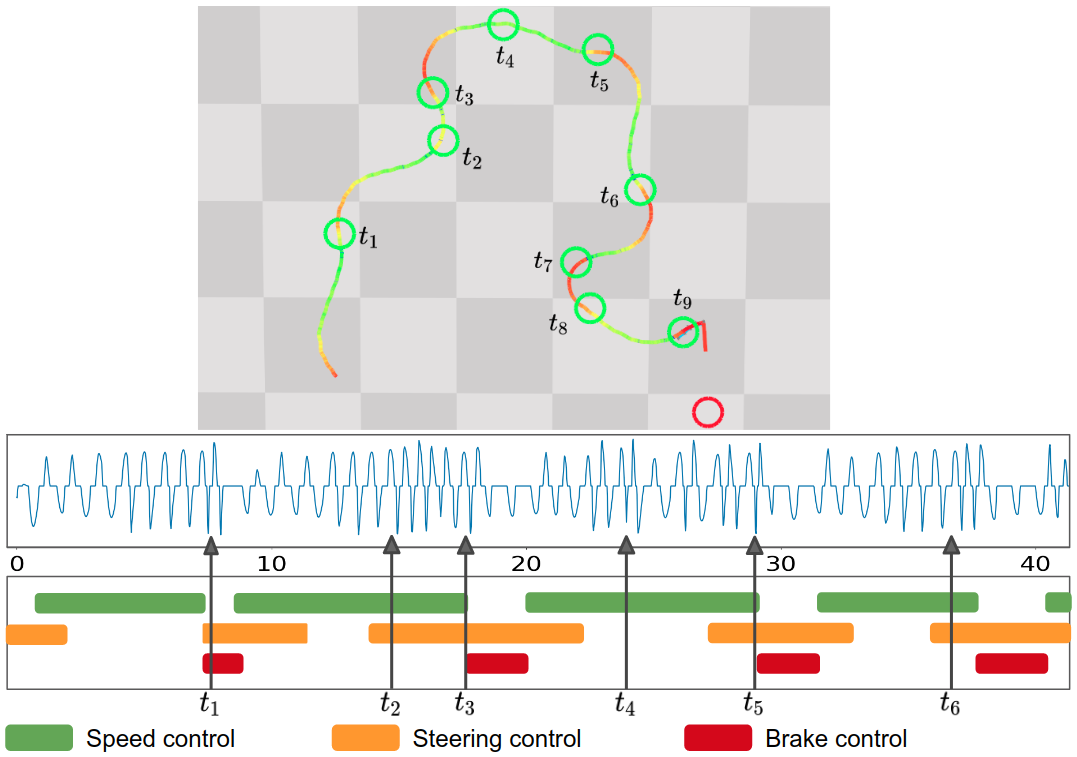}
    \caption{A sample trajectory and the corresponding control events on the first link \ac{cpg} output $\psi_1$.}
    \label{fig:trajresult}
    \vspace{-3ex}
\end{figure}



Figure~\ref{fig:trajresult} shows a sampled trajectory in the simulated snake robot controlled by the learned PPOC-CPG policy. Below the trajectory plot in Fig.~\ref{fig:trajresult} is the recorded pressure input trajectory to the first chamber. In the picture, green circles indicate the  goals reached successfully, and the red circle represents a new goal to be reached next. The colors on the path show the reward of the snake state, with a color gradient from red to green, indicating the reward value from low to high. Several maneuvering behaviors discussed in Section~\ref{sec:dis} are exhibited by the policy. First, as the higher frequency can result in lower locomotion speed, the trained policy presents a specific two-phase behavior --- (1) The  robot starts from a low oscillation frequency to achieve a high speed when it is far from the goal; (2) then it switches to  higher oscillation frequency using different options when it is getting closer to the goal. This allows it to stay close on the moving direction straight towards the goal. If this still does not work,  the tonic inputs will be used to force stopping the oscillation with the whole snake bending to the desired direction (see \ref{sec:dis}), and then restart the oscillation to acquire a larger turning angle.







\subsection{Experiments with the real robot}


\begin{figure}[h!]
\centering
\vspace{0.75em}
\begin{subfigure}[b]{0.9\columnwidth} 
\includegraphics[width=0.9\columnwidth]{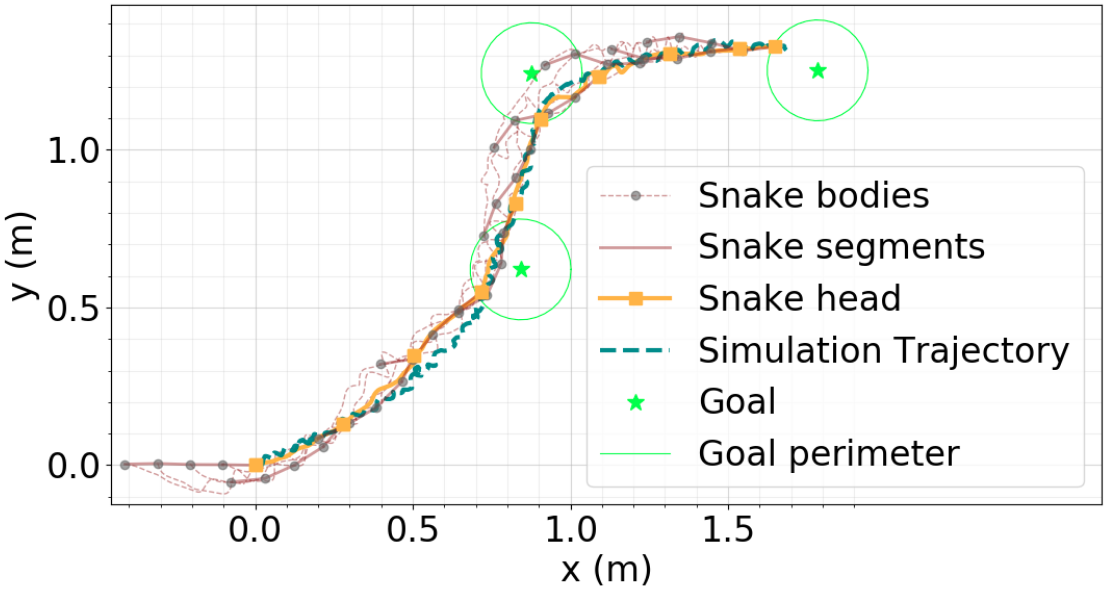} 
\caption{
\label{fig:simreal}}
\end{subfigure}

\hspace{-0.02\textwidth}
\begin{subfigure}[b]{0.9\columnwidth} 
\includegraphics[width=0.9\columnwidth]{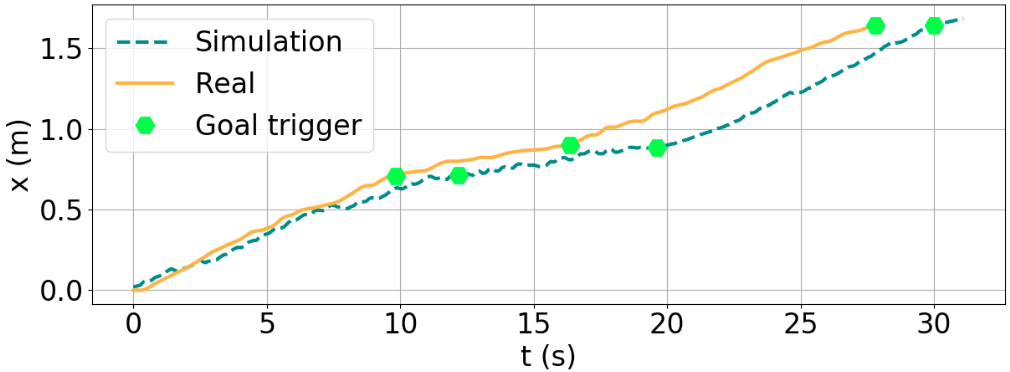}
\caption{
\label{fig:simrealx}}
\end{subfigure}

\begin{subfigure}[b]{0.9\columnwidth} 
\includegraphics[width=0.9\columnwidth]{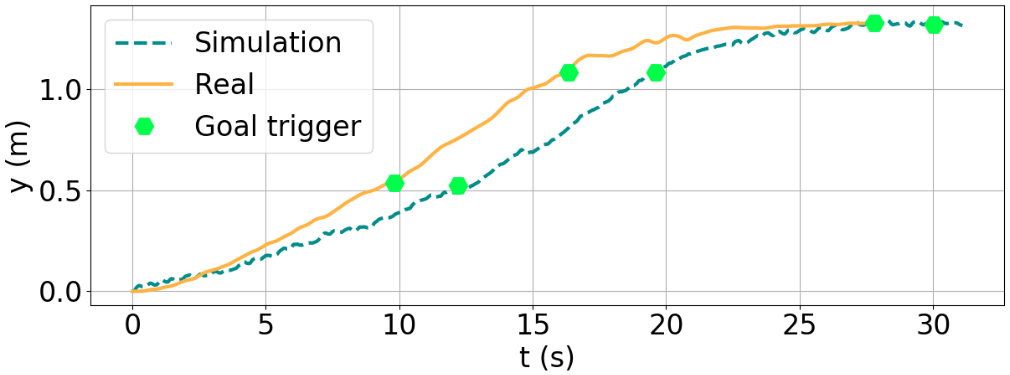}
\caption{
\label{fig:simrealy}}
\end{subfigure}

\caption{Head trajectory of the simulation and real snake robot in consecutive goal-reaching tasks using PPOC-CPG controller trained in simulation in (a) $x$-$y$ plane, (b) $x$-$t$ plane and (c) $y$-$t$ plane.}
\vspace{-2ex}
\label{fig:simrealresult}
\end{figure}

We apply the learned policy directly to the real robot. The policy is tested on goal-reaching tasks guided by a mobile robot with an accuracy radius of $r=0.175$ meter (The robot base has a $0.16$ meter radius). The learning policy obtains the mobile robot position using the Mocap system in $60 Hz$, and send the control commands to the robot through a low-latency wireless transmitter. A pair of example trajectories obtained from both simulation and real robot with identical policy are shown in Fig.~\ref{fig:simreal}. The real robot tracks the goals with an average speed of $0.092 \text{m/s}$, while the simulation robot with tuned contact friction (see Table~\ref{tab:dr} in Appendix) reaches an average speed of $0.083 \text{m/s}$. Though trained on fixed goals only, the policy can also follow the slowly moving target in the test. From Fig.~\ref{fig:simrealx} and \ref{fig:simrealy}, we notice that a delay of state evolution still exists between real robot and simulated robot, which is probably due to the inaccurate contact dynamics modeling by the physical engine. This could be fixed by either reducing the sim-to-real gap with more accurate modeling of the dynamics, or incorporating adaptive mechanism with velocity feedback to reduce the difference. We will leave this challenge as part of our future work. 

\section{Conclusion}
\label{sec:conclusion}
The contribution of this paper is two folds: First, we investigate the properties of Matsuoka oscillator for learning diverse locomotion skills in a soft snake robot. Second, we construct a PPOC-CPG net that uses a \ac{cpg} net to actuate the soft snake robot, and a reinforcement learning algorithm to learn a closed-loop near-optimal control policy that utilizes different oscillation patterns in the \ac{cpg} net. This learning-based control method shows promising results on goal-reaching and tracking behaviors in soft snake robots. This control architecture may be extended to motion control of other robotic systems, including bipedal and soft manipulators. Our next step is to verify the scalability of the proposed control framework on the soft snake robot with more body segments and extend it to a three-dimensional soft snake robot and to realize more complex motion and force control in soft snake robots using distributed sensors and visual feedback.


\bibliographystyle{IEEEtran}
\bibliography{refs.bib}
\vspace{4em}

\appendix

\begin{table}[h]
    \centering
    \caption{Parameter Configuration of Matsuoka \ac{cpg} Net Controller for the Soft Snake Robot.}
    \label{tab:config}
    \scalebox{1.0}{
    \begin{tabular}{p{3cm}|p{2cm}|p{2cm}}
     \hline 
     \textbf{Parameters} & \textbf{Symbols} & \textbf{Values} \\  \hline
        Amplitude & $A$ & 4.6062\\ 
        $*$Self inhibition weight & $b$ & \textbf{10.0355} \\
        $*$Discharge rate & $\tau_r$ & \textbf{0.2888} \\
        $*$Adaptation rate & $\tau_a$ & \textbf{0.6639} \\
        Period ratio & $K_f$ & 1.0\\[1ex] 
        \hline
        Mutual inhibition weights 
         & $a_i$ & 2.0935 \\[1ex]
        \hline
        Coupling weights 
         & $w_{ij}$ & 8.8669 \\
         & $w_{ji}$ & 0.7844 \\ [1ex]
     \hline
    \end{tabular}
}\end{table}
\vspace{-1em}
\begin{table}[h]
    \centering
    \caption{Domain randomization parameters}
    \scalebox{1.0}{
    \begin{tabular}{c|c|c}
        \hline
        \textbf{Parameter} & \textbf{Low} & \textbf{High} \\
        \hline
        Ground friction coefficient & 0.1 & 1.5 \\
        Wheel friction coefficient& 0.05 & 0.10 \\
        Rigid body mass ($kg$)& 0.035 & 0.075 \\
        Tail mass ($kg$)& 0.065 & 0.085 \\
        Head mass ($kg$) & 0.075 & 0.125 \\
        Max link pressure ($psi$) & 5 & 12\\
        Gravity angle ($rad$) & -0.001 & 0.001\\
        \hline
    \end{tabular}}
    \label{tab:dr}
\end{table}

\end{document}